\documentclass[10pt,twocolumn,letterpaper]{article}

\usepackage[pagenumbers]{cvpr} 

\usepackage{graphicx}
\usepackage{amsmath}
\usepackage{amssymb}
\usepackage{booktabs}
\usepackage{multirow}
\usepackage[dvipsnames]{xcolor}
\usepackage{dsfont}
\usepackage{makecell}
\DeclareMathOperator*{\argmax}{arg\,max} 
\newcommand{\paragrapht}[1]{\noindent\textbf{#1}} 
\usepackage{amsfonts}
\usepackage{amsmath}
\usepackage{algorithm}
\usepackage[noend]{algpseudocode}
\usepackage[dvipsnames]{xcolor}
\usepackage{bm}
\usepackage{placeins}
\usepackage[beginLComment=//~,endLComment=]{algpseudocodex}
\usepackage{ragged2e}

\newcommand{\trueblue}[1]{{\color{blue}{#1}}}
\newcommand{\truered}[1]{{\color{red}{#1}}}
\newcommand{\truegreen}[1]{{\color{ForestGreen}{#1}}}

\definecolor{cvprblue}{rgb}{0.21,0.49,0.74}
\usepackage[pagebackref,breaklinks,colorlinks,allcolors=cvprblue]{hyperref}

\def\paperID{*****} 
\def\confName{CVPR}
\def\confYear{2026}

\title{SpikeMatch: Semi-Supervised Learning with Temporal Dynamics \\ of Spiking Neural Networks}
\author{
Jini Yang$^{1,2}$ \qquad Beomseok Oh$^{2}$ \qquad Seungryong Kim$^{1}$ \qquad Sunok Kim$^{2\dagger}$ \\[5pt]
 \qquad $^{1}$KAIST AI \qquad $^{2}$Korea Aerospace University \\[5pt]
{\tt \href{https://cvlab-kaist.github.io/SpikeMatch}{https://cvlab-kaist.github.io/SpikeMatch}}
}

\begin{document}
\maketitle

{\let\thefootnote\relax\footnotetext{\hspace{-0.2cm}$^{\dagger}$Corresponding author.}}

\begin{abstract}
Spiking neural networks (SNNs) have recently been attracting significant attention for their biological plausibility and energy efficiency, but semi-supervised learning (SSL) methods for SNN-based models remain underexplored compared to those for artificial neural networks (ANNs). In this paper, we introduce SpikeMatch, the first SSL framework for SNNs that leverages the temporal dynamics through the leakage factor of SNNs for diverse pseudo-labeling within a co-training framework. By utilizing agreement among multiple predictions from a single SNN, SpikeMatch generates reliable pseudo-labels from weakly-augmented unlabeled samples to train on strongly-augmented ones, effectively mitigating confirmation bias by capturing discriminative features with limited labels. Experiments show that SpikeMatch outperforms existing SSL methods adapted to SNN backbones across various standard benchmarks.
\end{abstract}    
\section{Introduction}
\label{sec:intro}

\begin{figure}[t!]
	\centering
	{\includegraphics[width=\columnwidth]{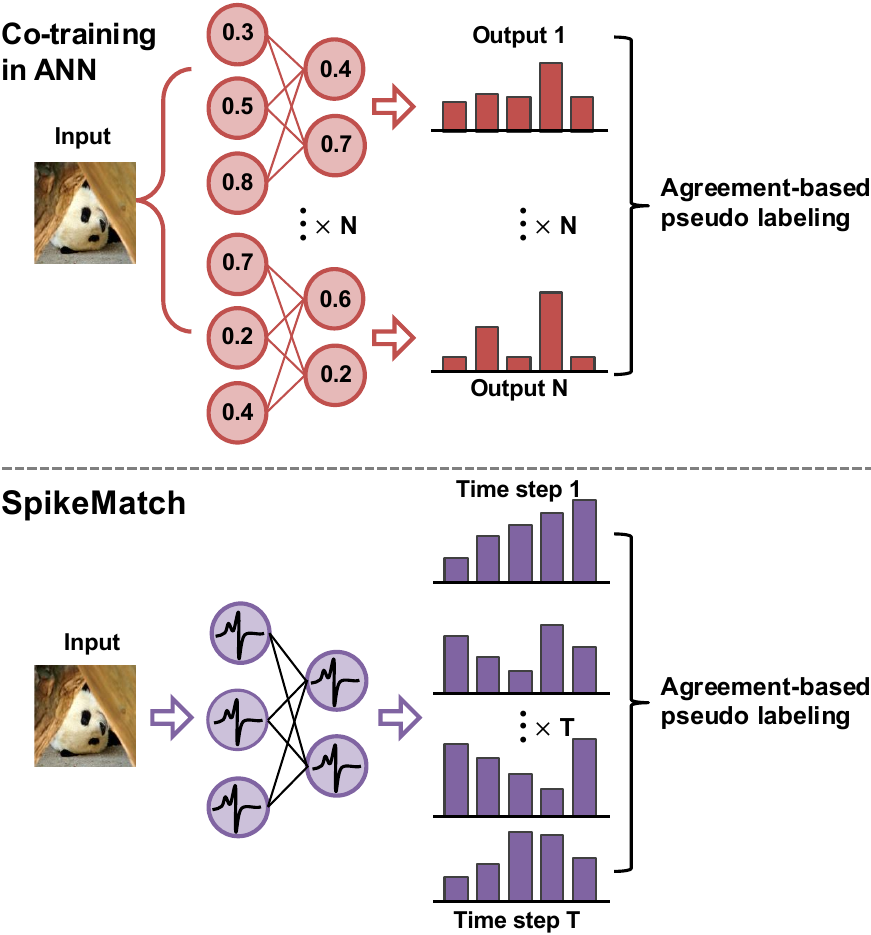}}
 	\caption{\textbf{Illustration comparing SNN-based SpikeMatch with ANN-based co-training.} As SNN-based training generates multiple predictions across time steps through binary spike signals, our approach leverages this temporal dynamic in predictions for agreement-based pseudo-labeling (\textbf{bottom}), in contrast to traditional ANN-based networks, which require additional parameters to produce diverse predictions (\textbf{top}).}\vspace{5pt}
	\label{fig:snn_diverse}
\end{figure}

Spiking neural networks (SNNs) have recently attracted growing interest for their biological plausibility and energy efficiency. Unlike artificial neural networks (ANNs), SNNs communicate through discrete spikes generated when the membrane potential exceeds a threshold. This mechanism introduces intrinsic temporal dynamics, as the membrane potential evolves over time. The leakage factor in leaky integrate-and-fire (LIF) neurons determines how much past information is retained, resulting in time-varying outputs even from static inputs. This temporal variability has led researchers to interpret SNNs as implicit ensemble models~\cite{dong2024temporal, zhao2024improving}, and they have been successfully applied to various computer vision tasks~\cite{fang2021deep, shi2024spikingresformer}.

Despite these successes, current approaches typically rely on abundant labeled data, which can be costly and labor-intensive to acquire. Thus, it is needed to develop suitable SNN method to address the challenge of label scarcity.
Semi-supervised learning (SSL) is one of the powerful solutions to this issue, as it leverages large amounts of unlabeled data in combination with a small set of labeled data to improve model generalization.

Previous studies~\cite{furuya2021semi, zhao2023semi, lee2023semi} used spike-timing-dependent plasticity (STDP) based two-stage SSL for SNNs, but these approaches struggle with deep networks and underperform compared to direct training. More recent work~\cite{nguyen2023semi} applies curriculum learning, outperforming STDP-based methods but remains limited to shallow networks and simple digit tasks without comparison to modern SSL algorithms.

On the other hand, ANNs have benefited from well-established semi-supervised learning techniques, particularly self-training with pseudo-labels~\cite{lee2013pseudo, cascante2021curriculum}. While effective, these methods suffer from confirmation bias, where early errors propagate over time and degrade performance. To address this, recent methods combine pseudo-labeling with consistency regularization. FixMatch~\cite{sohn2020fixmatch} introduces a confidence threshold and applies consistency between weak and strong augmentations. Building on this framework, recent methods further improve performance~\cite{zhang2021flexmatch, wang2022freematch, chen2023softmatch}, and recent theoretical work~\cite{li2024towards} emphasizes the importance of learning semantic variations through augmentation.

Another approach is co-training~\cite{blum1998combining}, originally developed for multi-view settings, has also been adapted to enhance prediction diversity in single-view scenarios. Recent studies use multiple classification heads or differently initialized models to mitigate confirmation bias through cross-labeling~\cite{chen2022semi}. Li et al.~\cite{li2023diverse} show that diversity is a critical factor in co-training, often outperforming FixMatch. However, these methods typically require extra computational cost from multiple initializations or architectural modifications~\cite{ke2019dual, fan2022ucc}.

In this paper, we leverage the temporal dynamics of SNNs within a co-training framework without requiring additional training resources. By simply increasing the leakage factor, we promote the SNN model to produce more diverse predictions and utilize this diversity for co-training through agreement-based pseudo-labeling.

With this approach, each temporal output of the SNN is trained using multiple perspectives, preventing extreme or dominant inaccurate pseudo-labels from being directly selected, thereby mitigating confirmation bias. Additionally, we incorporate successful SSL techniques from ANNs, such as distribution alignment and weak and strong augmentation, demonstrating their effectiveness in SNNs through experiments.

Our contributions are as follows:
\begin{itemize}
    \item We propose an agreement-based pseudo-labeling method that treats the temporal outputs of an SNN model as multiple predictions within a co-training framework. This approach addresses common challenges in semi-supervised learning, such as confirmation bias, by effectively leveraging distinct semantic features.
    \item We show that a higher leakage factor in the Leaky Integrate-and-Fire (LIF) neuron enhances prediction diversity, enabling the model to capture a more diverse set of discriminative features.
    \item We evaluate the proposed methods against other state-of-the-art semi-supervised learning methods used with ANNs by implementing them with an SNN-based backbone on the CIFAR, STL, and ImageNet datasets.
\end{itemize}

\section{Related Work}
\label{sec:rel_works}

\paragrapht{Semi-supervised learning.}
Semi-supervised learning (SSL) leverages large amounts of unlabeled data with limited labeled data. Two primary approaches are consistency regularization, which encourages consistent predictions under perturbations like augmentation~\cite{berthelot2019mixmatch}, dropout~\cite{sajjadi2016regularization}, or random max-pooling~\cite{sajjadi2016regularization}, and pseudo-labeling, which assigns pseudo-labels to unlabeled samples for supervised training~\cite{lee2013pseudo}.

Recent approaches combine both methods for improved performance. ReMixMatch~\cite{berthelot2019remixmatch} uses multiple augmentations and distribution alignment to address class imbalance issues of pseudo-labels, but using multiple strong augmentations can yield inconsistent predictions. FixMatch~\cite{sohn2020fixmatch} introduces a fixed threshold for confident pseudo-labeling using weak and strong augmentations, leading to a widely adopted confidence-aware framework~\cite{zhang2021flexmatch, berthelot2021adamatch, huang2023flatmatch, sosea2023marginmatch}. 
Motivated by FixMatch, FreeMatch~\cite{wang2022freematch} uses self-adaptive thresholding, considering both dataset-specific and class-specific factors, while SoftMatch~\cite{chen2023softmatch} proposes a Gaussian function to adjust the threshold, balancing the trade-off between pseudo-label quality and quantity. Recently, Li et al.~\cite{li2024towards} theoretically proved that the success of FixMatch-like methods comes from strong augmentation, which encourages the model to learn multiple feature views.

Co-training is another SSL strategy that originally leverages mutually exclusive features extracted from multiple views and distinct networks~\cite{blum1998combining}. However, since many datasets do not consist of multiple views, recent single-view co-training methods encourage diverse rather than independent features~\cite{wang2017theoretical}. Li et al.~\cite{li2023diverse} employ different architectures and augmentations to promote diversity and demonstrate that co-training outperforms FixMatch~\cite{sohn2020fixmatch} under various conditions in segmentation tasks. Multi-head co-training~\cite{chen2022semi} utilizes multiple classification heads and stochastic augmentation to generate diverse predictions for pseudo-labeling. JointMatch~\cite{zou2023jointmatch} trains two differently initialized networks for cross-pseudo-labeling, avoiding direct error accumulation. Similarly, UCC~\cite{fan2022ucc} employs two heads for cross-labeling in co-training to improve generalization ability for semi-supervised segmentation.

\paragrapht{Direct training of spiking neural networks.}
Spiking neural networks (SNNs) use the Heaviside step function as the non-linear activation of leaky integrate-and-fire (LIF) neurons, which receive continuous inputs, accumulate membrane potential with leakage, and generate binary spikes when the threshold is exceeded at each time step, resulting in dynamic properties~\cite{chowdhury2021towards}.

With these temporal characteristics, several studies have explored the unique information propagation in SNNs. Kim et al.~\cite{kim2023exploring} analyze Fisher information concentration, showing that the first few time steps exhibit similar characteristics and are designated as early time steps, while the last few time steps share similar traits and are classified as late time steps. Deng et al.~\cite{deng2022temporal} optimize the outputs of each time step separately for generalization ability. Furthermore, Dong et al.~\cite{dong2024temporal} leverage the distinct output distributions at each time step as an ensemble-modeled prediction for self-distillation training, whereas Zhao et al.~\cite{zhao2024improving} enforce consistency across these output differences to improve classification performance. More recently, Ding et al.~\cite{ding2025rethinking} investigate the differences in membrane potential at each time step and apply smoothing methods to enhance consistency.

Despite these advances in achieving high accuracy in image classification, most direct training methods focus on scenarios with abundant labeled data. Recently, some research has explored algorithms for direct training SNNs in label-scarce settings. For example, augmentation-based self-supervised learning to SNNs is applied in~\cite{bahariasl2024self}, but only achieved success on small datasets. Additionally, Nguyen et al.~\cite{nguyen2023semi} implemented pseudo-labeling for semi-supervised learning but tested it only on simple digit-recognition tasks. Extending this method to deeper SNNs for complex tasks substantially increases computational cost and training time due to the use of Monte-Carlo dropout for pseudo-labeling. Moreover, it relies on early firing time for pseudo-labeling, which becomes challenging in deeper networks due to their complex temporal dynamics. This makes pseudo-label generation less reliable and increases the risk of confirmation bias.
\section{Preliminaries}
\label{sec:method}
\paragrapht{Leaky integrate-and-fire model.}
We utilize the leaky integrate-and-fire (LIF) model~\cite{burkitt2006review} for SNN training. In the LIF model, the membrane potential at time step $t \in \{1, ..., T\}$ and layer $l$ before spiking, denoted by ${u}^{l,pre}(t)$, changes when the pre-synaptic input $W^l s^{l-1}(t)$ is injected as follows:
\begin{equation}
    {u}^{l,pre}(t) = \tau{u}^{l}(t-1) + W^l s^{l-1}(t),
    \label{eq:mem_update}
\end{equation}
where $\tau$ represents the leakage factor ranging within $(0,1)$ and ${u}^{l}(t-1)$ denotes the membrane potential after spiking at $t-1$. The pre-synaptic input $W^l s^{l-1}(t)$ is calculated by multiplying the synaptic weight ${W}^{l}$ by the input at time step  $t$, ${s}^{l-1}(t)$. The neuron generates spike output $s^{l}(t)$, when the ${u}^{l,pre}(t)$ exceed the firing threshold $V_{th}$:
\begin{equation}
    {s}^{l}(t) = \Theta({u}^{l,pre}(t) - V_{th}) = 
    \begin{cases}
    1,\quad if \ {u}^{l,pre}(t) \geq V_{th} \\
    0,\quad otherwise 
    \end{cases}
\end{equation}
\begin{equation}
    {u}^{l}(t) = {u}^{l,pre}(t)(1-{s}^{l}(t)),
    \label{eq:mem_reset}
\end{equation}
where $\Theta$ represents the Heaviside step function, and ${u}^{l}(t)$ is reset to zero if ${s}^{l}(t)$ is 1. In SNNs, this process is applied iteratively. Let $O^{t}(x)$ define the integrated pre-synaptic input $W^l s^{l-1}(t)$ of the last output layer for input $x$ without triggering a spike~\cite{rathi2021diet}. $p_{t}(y\mid x)= \mathrm{softmax}(O^{t}(x))$ denote the predicted class distribution produced by applying the softmax function to $O^{t}(x)$. This distribution $p_{t}(y \mid x)$ is used for classification~\cite{deng2022temporal}. 

\paragrapht{Leakage factor of LIF neuron.}
In~\cite{allen2020towards}, an ensemble model leveraging diversity solely through different random seeds improves performance by aggregating individual learned semantic views, enabling the model to capture multiple discriminative features. Motivated by this observation, we employ a different approach from existing ensemble-like SNN methods~\cite{zhao2024improving,ding2025rethinking}, which typically utilize temporal dynamics with abundant data by enforcing consistency. Instead, we leverage this diversity to learn multiple distinctive views in a limited-data setting with a single spiking network. We achieve this by using a higher the leakage factor unlike ANN require additional training or multi-view data. To analyze the role of the leakage factor in membrane potential dynamics, we assume that neurons do not emit spikes between time steps. From Equation~\eqref{eq:mem_reset}, we can expand this recursively as for  $t\in [2, T)$, where  $t^{\prime} \in [1, t)$ ~\cite{ding2022snn}:

\begin{equation}
u^l(t)=
\tau^{t-t^\prime}u^l(t^\prime) +  \sum_{i=t^{\prime}+1}^{t} \tau^{t - i} W^l s^{l-1}(i),
\end{equation}

To quantify the diversity of membrane potential, we select a time step $t^{\prime} \in [1, t)$ and subtract its membrane potential from $u^l(t)$:
\begin{equation}
\begin{aligned}
\vert u^l(t) - &u^l(t^{\prime}) \vert= \\&
\vert (\tau^{t-t^\prime}-1)u^l(t^\prime) + \sum_{i=t^{\prime}+1}^{t} \tau^{t - i} W^l s^{l-1}(i) \vert.
\end{aligned}
\end{equation}
For larger $\tau$, past spike information persists for a longer duration, increasing the difference between time steps. This enhances temporal diversity, providing the model with more opportunities to learn distinct features in the agreement-based co-training framework. Based on this property, we design our SSL method to leverage the inherent temporal dynamics of SNNs, enabling effective pseudo-labeling without additional diversity constraints.

\section{SpikeMatch}
\begin{figure*}[t!]
	\centering
	\includegraphics[width=\textwidth]{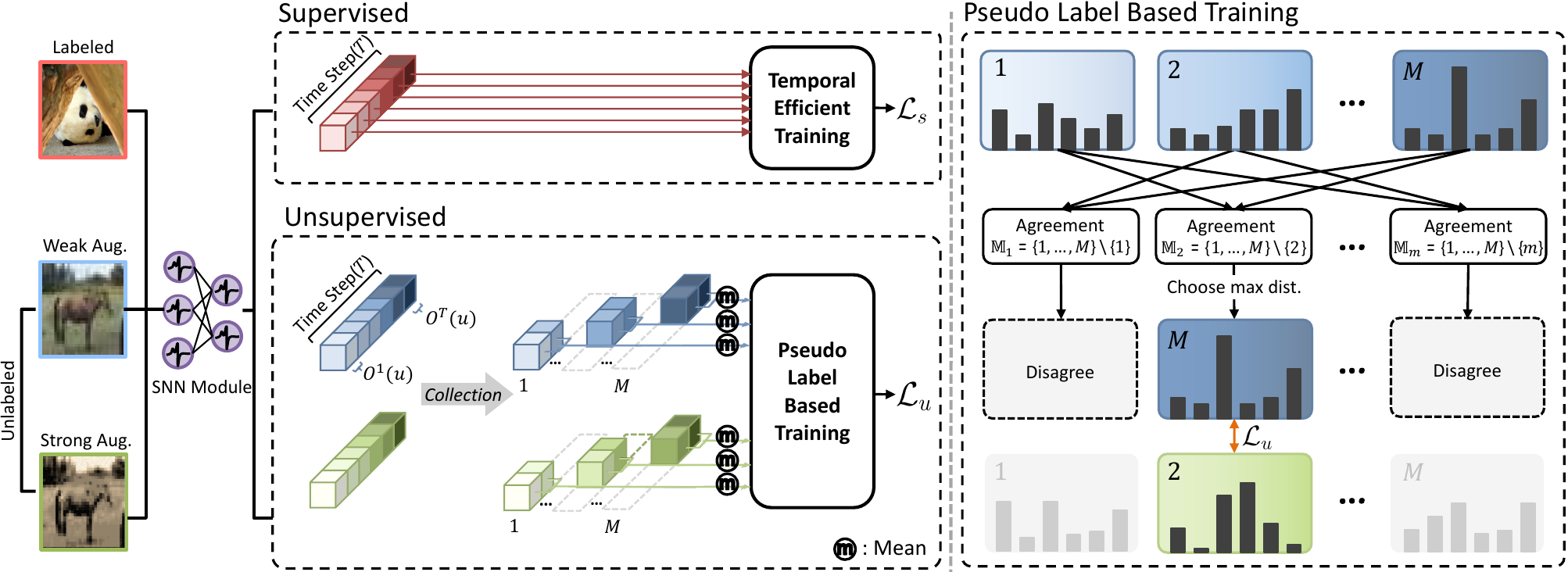}
 	\caption{\textbf{An Overview of SpikeMatch (left) and pseudo-label based training (right).} \textbf{(left)}: The weakly augmented labeled image (\truered{red}) produces $T$ outputs through the SNN model and optimized by TET loss. For unlabeled data, both weakly (\trueblue{blue}) and strongly (\truegreen{green}) augmented images pass through the same SNN model. The $T$ outputs produced by each image are partitioned and averaged into $M$ predictions. The predictions from the weakly augmented image serve as the soft pseudo-label, using their agreement to optimize the strongly augmented image via cross-entropy. \textbf{(right)}: A pseudo-label from one collection (\truegreen{green}) is used when the remaining $M-1$ collections (\trueblue{blue}) agree on the same class with the highest probability, and the prediction with the highest confidence is selected as the soft pseudo-label.}
	\label{fig:overall_framework}
\end{figure*}

\subsection{Problem statement and motivation}
In this section, we introduce our detailed methods utilizing an SNN-based semi-supervised learning (SSL) approach that leverages co-training inspired by the temporal dynamics of SNNs. 
For a $C$-class classification problem, let a batch of $B$  labeled samples be denoted as $\mathcal{X}=\{(x_b, y_b):b \in (1,...,B)\}$ and a batch of $\mu B$ unlabeled samples as  $\mathcal{U}=\{u_b:b \in (1,...,\mu B)\}$, where $\mu$ is a hyperparameter representing the relative size of unlabeled samples $\mathcal{U}$ to labeled samples $\mathcal{X}$, and $y_b$ is the one-hot label. 

Unlike traditional ANNs, which process static images spatially, SNNs propagate information through both spatial and temporal dimensions, producing multiple predictions over time, as shown in Figure~\ref{fig:snn_diverse}. Based on these distinctive characteristics of the SNN model, we propose a novel SNN-based SSL model that combines the temporal dynamics of spiking neural networks with a co-training mechanism to generate agreement-based pseudo-labels, utilizing consistency regularization. The overall architecture is visualized in Figure~\ref{fig:overall_framework}.

\begin{algorithm*}[h]
\caption{SpikeMatch algorithm}
\begin{algorithmic}[1]
\Require Labeled batch $\mathcal{X}=\{(x_b, y_b):b \in (1,...,B)\}$, unlabeled batch $\mathcal{U}=\{u_b:b \in (1,...,\mu B)\}$, time step $T$, number of collected predictions $M$, unsupervised loss weight $\lambda$, weak and strong augmentations $\alpha(\cdot)$, $\mathcal{A}(\cdot)$
\LComment{Supervised learning on $\mathcal{X}$}
\State $\mathcal{L}_{s} = \frac{1}{BT}\sum^{B}_{b=1}\sum^{T}_{t=1} \text{H}(y_b, p_{t}(y \mid \alpha(x_b)))$ \quad\quad\quad\quad\quad\quad\quad\quad\quad\quad\quad\Comment{Compute TET loss for labeled samples}
\LComment{Unsupervised learning on $\mathcal{U}$}
\For{$b = 1$ to $\mu B$}
    \For{$m = 1$ to $M$} \quad\quad\quad\quad\quad\quad\quad\quad\quad\quad\Comment{Calculate the class distributions for $m$}
        \State $q^m_b = \frac{1}{|t_m|}\sum_{t \in t_m} O^{t}(\alpha(u_{b}))$ 
        \State $g_{m}(y \mid \mathcal{A}(u_{b})) = \frac{1}{|t_m|}\sum_{t \in t_m} O^{t}(\mathcal{A}(u_{b}))$
    \EndFor
    \For{$m = 1$ to $M$} \quad\quad\quad\quad\quad\quad\quad\quad\quad\Comment{Select the confident pseudo-label}
        \State $\hat{q}^{m}_{b} =q^{i}_{b},\quad \mathrm{where}\quad i=\argmax_{j \in \mathbb{M}_{m}}
    \left( \max_c \, g_{j}(y=c \mid \alpha(u_{b})) \right)$
    \EndFor
    \For{$m = 1$ to $M$} \quad\quad\quad\quad\quad\quad\quad\quad\Comment{Generate pseudo-label with agreement}
        \State Set $\hat{c}^m=\argmax_{c}g_{i}(y=c \mid \alpha(u_{b}))$ for any $i \in \mathbb{M}_{m}$ 
        \State $\ell_{b}=\sum^ {M}_{m=1}\prod\limits_{i} \mathds{1} ( \argmax_c g_{i}(y=c \mid \alpha(u_{b})) = \hat c^m )\times \text{H}(\hat{q}^{m}_{b},g_{m}(y \mid \mathcal{A}(u_{b})))$, \quad $i \in \mathbb{M}_{m}$
    \EndFor
\EndFor
\State $\mathcal{L}_{u} = \frac{1}{\mu B}\sum^{\mu B}_{b=1} \ell_b$
\State return $\mathcal{L}_{s}$ and $\mathcal{L}_{u}$
\end{algorithmic}
\end{algorithm*}

\subsection{Temporal efficient supervised learning}
In SSL, the amount of labeled data $|\mathcal{X}|$ is comparatively small compared to the unlabeled data $|\mathcal{U}|$, which can degrade the generalization ability of the model~\cite{huang2023flatmatch}. We address this problem by leveraging the temporal dynamics of SNNs without pre-training or adding extra parameters. For supervised learning, the loss from labeled data is defined by the Temporal Efficient Training (TET) loss~\cite{deng2022temporal}, which is known to enhance the generalization ability of the model by optimizing the output distribution at each time step $p_{t}(y\mid{x})$ individually. The supervised loss $\mathcal{L}_{s}$ is defined as:
\begin{equation}
    \mathcal{L}_{s} =  \frac{1}{BT}\sum^{B}_{b=1}\sum^{T}_{t=1} \text{H}(y_b, p_{t}(y \mid \alpha(x_b))),
\end{equation}
where $\text{H}\left( \cdot \right)$  denotes the cross-entropy loss and  $\alpha\left( \cdot \right)$ represents a weak augmentation.

\subsection{Agreement-based pseudo-labeling}
\paragrapht{Pseudo-labeling.} With limited labeled data, consistency regularization with weak and strong augmentations effectively helps to avoid overfitting~\cite{sohn2020fixmatch}. Following this approach, we incorporate consistency regularization along with pseudo-labeling. However, using pseudo-labels from a single prediction can lead to error accumulation~\cite{feng2022dmt}. To address this challenge, motivated by co-training~\cite{chen2022semi, zou2023jointmatch}, our method for unsupervised training focuses on generating complementary predictions through temporal variations, treating these variations as distinct output distributions, as visualized in Figure~\ref{fig:overall_framework}.

In~\cite{kim2023exploring}, the authors investigate the temporal information in SNNs using Fisher information concentration and claim that early and late time steps focus on different feature information.
Since pseudo-labeling with agreement across predictions at every time step is time-consuming, for simplicity and diversity in predictions, we reduce the total $T$ temporal outputs to $M$ by sequentially grouping every $T/M$ outputs across the time steps $\{1, ..., T\}$. By averaging each of these groups, we obtain $M$ predictions, denoted as $g_{m}$ where $m \in \{1, \dots, M\}$, which serve as the final set of predictions for unsupervised training:
\begin{equation}
 g_{m}(y \mid u_{b})=\frac{1}{|t_m|}\sum_{t \in t_m} O^{t}(u_{b}),
\end{equation}
where $t_m = \{ t \mid (m - 1) \cdot \frac{T}{M} + 1 \leq t \leq m \cdot \frac{T}{M} \}$, $|t_m|=\frac{T}{M}$.
Each temporal output $O^{t}(u_{b})$ over $T$ is divided into $M$  collections and averaged, resulting in $M$ predicted class distributions $g_{m}(y \mid u_{b})$, where  $m \in \{1, \dots, M\}$. In our method, these predictions interact with others through pseudo-labels on unlabeled data, leveraging temporal prediction diversity to generate more diverse pseudo-labels. 
This partitioning approach, similar to other co-training methods~\cite{zou2023jointmatch, fan2022ucc, chen2022semi, li2023diverse}, also demonstrates robustness to confirmation bias that can arise from incorrect predictions in early training stages.

\begin{table*}[hbt!]
\centering
    \resizebox{\linewidth}{!}
    {
    \begin{tabular}{l|cccccc}
    \toprule[1pt]
    Dataset & \multicolumn{3}{c}{CIFAR-10} & \multicolumn{3}{c}{CIFAR-100} \\ 
    \cmidrule(lr){1-1} \cmidrule(lr){2-4} \cmidrule(lr){5-7} 
    Labels &4 &25 &400 &4 &25 &100\\
    \cmidrule(lr){1-1} \cmidrule(lr){2-2} \cmidrule(lr){3-3} \cmidrule(lr){4-4} \cmidrule(lr){5-5} \cmidrule(lr){6-6} \cmidrule(lr){7-7}
    UDA~\cite{xie2020unsupervised}        &$69.72_{\pm6.65}$ &$87.76_{\pm 0.21}$ &$90.06_{\pm 0.06}$ &$44.66_{\pm 1.63}$ &$60.35_{\pm0.67}$ &$66.51_{\pm 0.15}$ \\ 
    FixMatch~\cite{sohn2020fixmatch}        &$67.97_{\pm4.42}$ &$86.84_{\pm0.91}$ &$89.40_{\pm0.23}$ &$38.39_{\pm0.91}$ &$58.40_{\pm0.48}$ &$64.75_{\pm0.17}$ \\ 
    AdaMatch~\cite{berthelot2021adamatch}          &$82.13_{\pm6.98}$ &$87.15_{\pm0.62}$ &$89.44_{\pm0.11}$ &$44.54_{\pm2.52}$ &$59.52_{\pm0.24}$ &$65.86_{\pm0.44}$  \\ 
    FreeMatch~\cite{wang2022freematch}          &$85.65_{\pm4.16}$ &$87.86_{\pm0.34}$ &$89.64_{\pm0.27}$ &$46.49_{\pm2.75}$ &$62.01_{\pm0.27}$ &$67.10_{\pm0.35}$  \\
    SoftMatch~\cite{chen2023softmatch}         &$84.27_{\pm7.69}$ &$87.97_{\pm0.18}$ &$89.91_{\pm0.28}$ &$45.99_{\pm3.06}$ &$61.97_{\pm0.53}$ &$67.35_{\pm0.08}$\\
    RegMixMatch~\cite{han2025regmixmatch} &$82.59_{\pm4.02}$ &$87.36_{\pm0.55}$ &$90.08_{\pm0.65}$ &$38.49_{\pm3.09}$ &$60.27_{\pm1.96}$ &$66.69_{\pm0.24}$ \\
    \cmidrule(lr){1-1} \cmidrule(lr){2-4} \cmidrule(lr){5-7} 
    SpikeMatch w/o TET   &${87.98}_{\pm0.51}$ &${88.13}_{\pm0.37}$ &$\textbf{91.00}_{\pm0.002}$ &$47.45_{\pm1.79}$ &${63.18}_{\pm0.09}$ &${69.37}_{\pm0.003}$\\
    SpikeMatch     &$\textbf{88.13}_{\pm0.79}$ &$\textbf{88.36}_{\pm0.06}$ &${90.93}_{\pm0.04}$ &$\textbf{47.52}_{\pm2.18}$ &$\textbf{63.34}_{\pm0.04}$ &$\textbf{69.50}_{\pm0.17}$\\
    \bottomrule[1pt]
    \end{tabular}
    }
    \caption{\textbf{Accuracy (\%) on CIFAR-10 and CIFAR-100 using 3 different random seeds.} The best results are in bold.}
    \label{tab:main}
\end{table*}

\begin{table}[]
    \centering
    \begin{tabular}{lcccc}
    \toprule[1pt]
        Seed &0 &1 &2 &Mean  \\
        \hline
        FreeMatch~\cite{wang2022freematch} &56.73 &49.94 &57.00 &$54.55_{\pm4.00}$ \\
        SoftMatch~\cite{chen2023softmatch} &58.40 &50.58 &58.14 &$55.70_{\pm4.44}$ \\
        \hline
        SpikeMatch &62.36 &64.69 &61.36 &$62.80_{\pm1.70}$ \\
         \bottomrule[1pt]
    \end{tabular}
    \caption{\textbf{Accuracy (\%) on STL-10.}}
    \label{tab:stl}
\end{table}
To facilitate interactions among predictions with consistency regularization, we denote  $q^{m}_{b} = g_{m}(y \mid \alpha(u_{b}))$ as the predicted class distribution for each $m$ in the $M$ predictions on weakly augmented unlabeled data.
To benefit from diversity and alleviate confirmation bias, we generate the pseudo-label $\hat{q}^m_b$ from each $q^{m}_{b}$ using the other $M-1$ predictions in $\mathbb{M}_{m}=\{1,...,M\}\setminus\{m\}$,
where $\mathbb{M}_{m}$ represents the set of indices excluding $m$. The pseudo-label $\hat{q}^m_b$ is defined as:
\begin{equation}
    \hat{q}^{m}_{b} =q^{i}_{b},\, \mathrm{where}\, i=\argmax_{j \in \mathbb{M}_{m}}
    \left( \max_c \, g_{j}(y=c \mid \alpha(u_{b})) \right),
\end{equation}
where $i$ indicates the index within the set $\mathbb{M}_{m}$ that yields the highest value of $g_{i}(y=c \mid \alpha(u_{b}))$ over the class $c$.
These pseudo-labels are trained with additional constraints based on the agreement of other predictions $q^i_b (i\in\mathbb{M}_m)$, utilizing temporal dynamics similar to co-training methods, which are known to reduce error accumulation~\cite{zou2023jointmatch, dong2018tri}. If the maximal classes among $q^i_b (i\in\mathbb{M}_m)$ are the same, the $\hat{q}^{m}_{b}$ is used as a pseudo-label for training with the output prediction on $m$ of the strongly augmented unlabeled data, $g_{m}(y \mid \mathcal{A}(u_{b}))$ as shown in Figure~\ref{fig:overall_framework} (right). This strategy effectively integrates co-training principles with the temporal nature of SNNs to enhance learning from both labeled and unlabeled data, while preserving prediction diversity and utilizing a large set of unlabeled samples in self-training.

\paragrapht{Loss function.} The cross-entropy loss on the unlabeled data is then calculated as follows:
\begin{equation}
    \begin{aligned}
        \mathcal{L}_{u} =\frac{1}{\mu B}\sum^{\mu B}_{b=1}\sum^ {M}_{m=1}\prod\limits_{i \in \mathbb{M}_{m}} \mathds{1} &( \argmax_c g_{i}(y=c \mid \alpha(u_{b})) = \hat c^{m} )\\\times
        & \text{H}(\hat{q}^{m}_{b},g_{m}(y \mid \mathcal{A}(u_{b}))),
    \end{aligned}
\end{equation}
where $\hat c^{m}$ denotes the class with the maximum probability in $q^i_b$, with $i$ representing any element in  $\mathbb{M}_{m}$, and $\mathcal{A}$ represents a strong augmentation.
Finally, the overall loss is written as:
\begin{equation}
    \mathcal{L} = \mathcal{L}_{s} + \lambda\mathcal{L}_{u}.
\end{equation}
where $\lambda$ is a hyper-parameter that balances the supervised and unsupervised loss.

\subsection{Multi-view feature learning}
Originally, co-training required multi-view data to achieve diversity. However, recent co-training methods generate diversity by using differently initialized models trained on the same dataset. These approaches have been shown to outperform FixMatch-based architectures in semi-supervised learning~\cite{zou2023jointmatch}. This diversity allows the model to learn distinct features in an ensemble-like manner~\cite{allen2020towards}. We leverage this diversity through the leakage factor of the LIF neuron, where increasing it generates more diverse predictions over time without requiring additional constraints for diversity.

Our agreement-based method leverages the co-training framework to effectively capture diverse features under limited labeled data. Unlike FixMatch-like methods that rely on confidence thresholding, our approach does not impose such constraints. In the early stages of training, extreme predictions from a single view are filtered out if they are not in agreement with others. This process prevents the temporal outputs from being influenced by a single extreme prediction, reducing their impact and improving robustness against confirmation bias. Additionally, each temporal output is optimized based on complementary outputs, providing an opportunity for the model to learn different features from other temporal predictions.

\section{Experiments}
\label{sec:experiments}

\subsection{Implementation details}
We evaluate our approach on CIFAR~\cite{krizhevsky2009learning}, STL-10~\cite{coates2011analysis}, and ImageNet~\cite{deng2009imagenet}, which are widely used benchmark datasets for semi-supervised image classification. Additional results including neuromorphic dataset are included in the Appendix. Since many SNN methods adopt VGG-based architectures~\cite{deng2022temporal}, we use VGG networks~\cite{simonyan2014very} for our main experiments. We set the number of time steps $T$ to 4 and generate $M = 3$ predictions for each unlabeled sample. When $T$ is not divisible by $M$, we divide the outputs into segments of sizes ${1, 1, 2}$ and average the outputs within each segment to produce the $M$ predictions. Further training details are provided in the Appendix.

\subsection{Main results}
\paragrapht{Analysis on CIFAR.} We evaluate SpikeMatch on CIFAR-10, CIFAR-100, and STL-10, comparing it against recent SSL methods originally designed for ANNs, including UDA~\cite{xie2020unsupervised}, FixMatch~\cite{sohn2020fixmatch}, AdaMatch~\cite{berthelot2021adamatch}, FreeMatch~\cite{wang2022freematch}, SoftMatch~\cite{chen2023softmatch}, and RegMixMatch~\cite{han2025regmixmatch}. To ensure a fair comparison, we re-implement these methods with an SNN backbone using their original settings. All results are averaged over three seeds with standard deviations reported.
As shown in Table~\ref{tab:main}, our approach achieves superior performance compared to other methods, including recent state-of-the-art methods FreeMatch, SoftMatch, and RegMixMatch, across various label regimes on CIFAR-10 and CIFAR-100.
Furthermore, SpikeMatch demonstrates a notably low standard deviation when using only four examples per class.

\begin{figure}[]
	\centering
       \includegraphics[width=\linewidth]{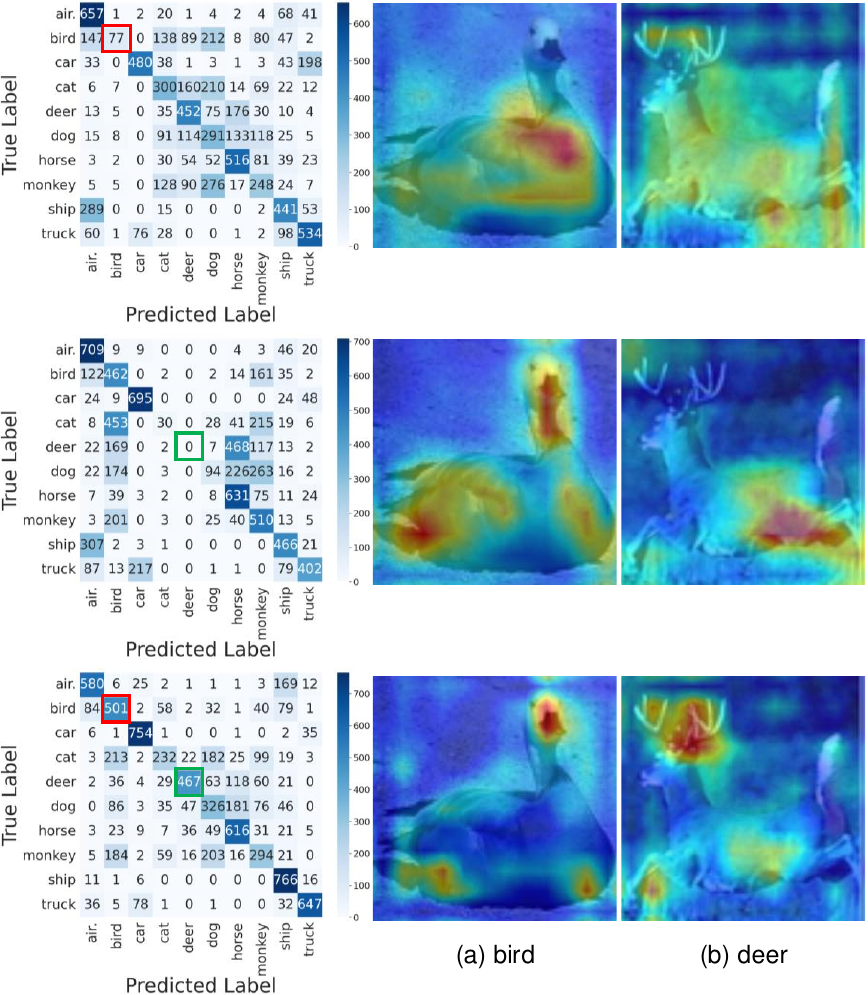}
    \caption{\textbf{Confusion matrix and SAM on STL-10.} Unlike FreeMatch (\textbf{first row}), which is confused by the bird class, and SoftMatch (\textbf{second row}), which struggles with the deer class, our method (\textbf{last row}) captures more discriminative features, such as the beak of (a) bird and the antlers of (b) deer. This results in a more diagonal confusion matrix, making the model less prone to confusion among similar classes.}
	\label{fig:confusion_stl10}
\end{figure}

\paragrapht{Analysis on STL-10.} We further evaluate our approach on the more challenging STL-10 dataset, which contains high-resolution examples. 
We compare our method against recent state-of-the-art approaches, FreeMatch and SoftMatch, using a randomly selected subset of 40 labeled samples. As shown in Table~\ref{tab:stl}, our method achieves 7.10\% higher accuracy than SoftMatch while maintaining a low standard deviation. For visualization, we utilize spiking activation map (SAM)~\cite{kim2021visual}, to generate temporal localization maps that highlight regions of interest. Figure~\ref{fig:confusion_stl10} illustrates that under scarce label conditions, directly adapting existing methods to SNNs often leads to class imbalance. These results suggest that our agreement-based temporal dynamic approach effectively captures diverse discriminative features, enhancing the distinction between frequently confused classes.
\begin{table}[t]
\centering
    \begin{tabular}{l|cc}
    \toprule[1pt]
        Methods &Top-1 &Top-5  \\
        \cmidrule(lr){1-1} \cmidrule(lr){2-2} \cmidrule(lr){3-3}
        SoftMatch~\cite{chen2023softmatch} &45.66 &71.44  \\
        SpikeMatch &\textbf{49.68} &\textbf{75.07} \\
         \bottomrule[1pt]
    \end{tabular}
    \caption{\textbf{Accuracy (\%) on ImageNet using 100 labels per class.}}\vspace{-5pt}
    \label{tab:imagenet}
\end{table}

\paragrapht{Results on ImageNet.}
To evaluate our method on a deeper network and a larger dataset, we conduct experiments using SEW-ResNet50~\cite{fang2021deep} on ImageNet~\cite{deng2009imagenet} with 100 labels per class. The results presented in Table~\ref{tab:imagenet} show that our method outperforms SoftMatch even with a deep backbone network on a challenging task.

\begin{table}[]
    \centering
    \begin{tabular}{l|cc}
    \toprule[1pt]
         Methods &Accuracy &Energy efficiency \\
        \cmidrule(lr){1-1} \cmidrule(lr){2-2} \cmidrule(lr){3-3}
        FixMatch (ANN) &$78.69_{\pm9.13}$ &$1\times$  \\
        SoftMatch (ANN) &$79.04_{\pm12.54}$ &$1\times$  \\
        \cmidrule(lr){1-1} \cmidrule(lr){2-2} \cmidrule(lr){3-3}
        SpikeMatch &\textbf{$81.92_{\pm3.06}$} &$20.8\times$ \\
         \bottomrule[1pt]
    \end{tabular}
    \caption{\textbf{Accuracy (\%) and energy consumption (including MAC) on CIFAR10 with only one label per class.}}
    \label{sup_tab:tab_energy_ann}
\end{table}


\paragrapht{Energy and label efficiency.}
As we leverage the inherent temporal dynamics of SNNs in a co-training manner, SpikeMatch achieves remarkable performance in the most challenging setting of CIFAR-10 with only one labeled sample per class as shown in Table~\ref{sup_tab:tab_energy_ann}. Unlike earlier setups that simply replace the backbone of existing methods with an SNN, this section compares against SoftMatch with its original ANN backbone. Despite this, SpikeMatch not only outperforms SoftMatch, a state-of-the-art ANN-based method, but also demonstrates the ability to extract highly discriminative features from unlabeled data under extreme label scarcity. Moreover, our method fully preserves the energy efficiency of SNNs, achieving over 20.8 times higher energy efficiency compared to the ANN-based counterpart. This makes SpikeMatch well suited for deployment in scenarios constrained by both energy and labeled data.

\begin{figure}[t]
    \centering
    {\includegraphics[width=0.44\textwidth]{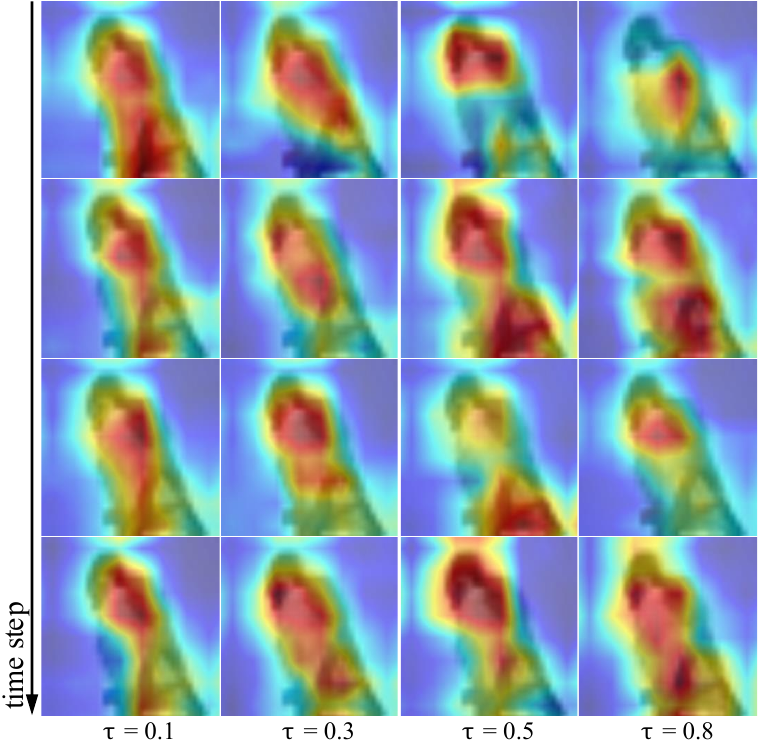}}
 	\caption{\textbf{SAM on leakage factor and time step variance.} As the leakage factor increases, SAM visualizations at each time step become more diverse, providing more opportunities for the model to learn discriminative features.}\vspace{-5pt}
	\label{fig:sam_diverse}
\end{figure}
\subsection{Configuration and ablation study}
\begin{figure}[t]
	\centering	{\includegraphics[width=\linewidth]{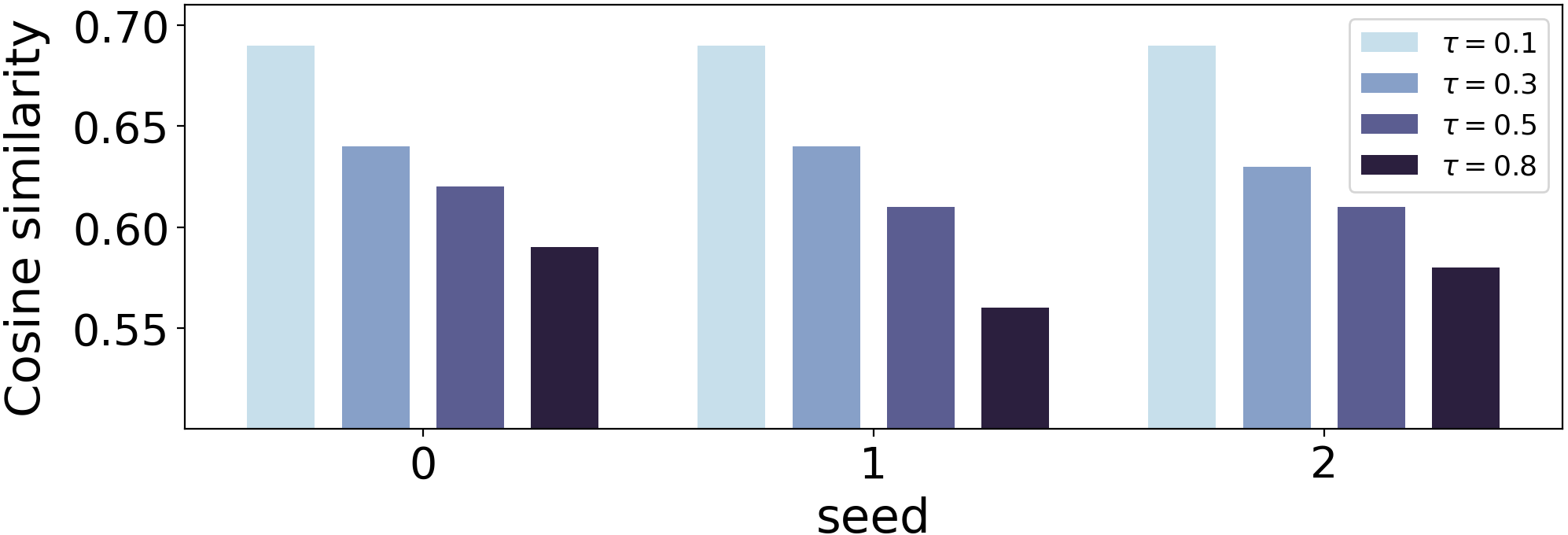}}
 	\caption{\textbf{Cosine similarity at varying leakage factors ($\tau$).} Each column represents the average of 6 cosine similarity values computed across 4 time steps for each $\tau$, at seeds 0, 1, and 2. The cosine similarity is calculated from 6 combinations of features from the 4 time steps at the last layer.}
	\label{sup_fig:cos_sim}
\end{figure}

\paragrapht{Leakage factor of LIF.}
To demonstrate that a higher leakage factor induces more diverse predictions over time, we conduct an ablation study by varying the leakage factor. As shown in Figure~\ref{fig:sam_diverse}, increasing the leakage factor results in more diverse SAM visualizations at last layer. Additionally, we calculate the cosine similarity for every pair of time steps from the last layer. As shown in Figure~\ref{sup_fig:cos_sim}, without any explicit diversity constraints, the mean cosine similarity at each leakage factor consistently decreases across all seeds. This indicates that the model produces more diverse temporal features as $\tau$ increases. 
This increased diversity allows the model to learn a wider range of features, ultimately improving SSL performance.  Additional empirical analyses of this diversity are provided in the appendix. 
As summarized in Table~\ref{tab:tau}, both SoftMatch and our method achieve higher performance with a higher leakage factor. However, the performance gap is more notable in our method. This is because our agreement-based approach is more effective at leveraging minor features that emerge from temporal diversity, leading to greater improvements in SSL performance. However, a higher leakage factor accumulates membrane potential and generally produces more spikes, leading to increased energy consumption. To balance efficiency and performance, we set $\tau$ to 0.5 by default.

\begin{table}[t]
    \centering
    \resizebox{1.00\columnwidth}{!}{
    \begin{tabular}{l|ccccc}
    \toprule[1pt]
        Leakage factor &0 &0.1 &0.3 &0.5 &0.8  \\
        \cmidrule(lr){1-1} \cmidrule(lr){2-2} \cmidrule(lr){3-3}         \cmidrule(lr){4-4} \cmidrule(lr){5-5} \cmidrule(lr){6-6}

        SoftMatch &79.71 &84.27 &84.41 &84.27 &85.52 \\
        SpikeMatch &80.16 &85.43 &86.33 &88.13 &88.17 \\
         \bottomrule[1pt]
    \end{tabular}}
    
    \caption{\textbf{Averaged accuracy (\%) over three seeds on CIFAR-10 with 4 labels per class under different leakage factors.}}\vspace{-5pt}
    \label{tab:tau}
\end{table}


\paragrapht{Ablation Study.} 
We evaluated the contributions of each technique in SpikeMatch through ablation studies on CIFAR with 4 labels per class, as shown in Table~\ref{tab:abl_module}. We compared SpikeMatch with an averaged prediction model averaging temporal outputs (single prediction model), an intra prediction model without agreement-based pseudo-labeling, a version without distribution alignment (DA)~\cite{berthelot2019remixmatch}, and one using a threshold as an additional constraint for pseudo-labeling.

The averaged prediction model showed sub-optimal results because it cannot leverage intrinsic diversity of SNN for pseudo-labeling. Using intra prediction also resulted in performance degradation, as it does not complement each prediction with distinct features. In contrast, SpikeMatch leverages cross-labeling to effectively utilize this diversity. To verify the DA in SNNs, we performed ablation experiments by applying SpikeMatch without DA. There was a performance drop because DA aligns the pseudo-labels from all collected predictions. We also added a threshold as an additional constraint in agreement-based pseudo-labeling but observed a performance decline, possibly because setting the threshold to 0.8 is not optimal for our SNN network. However, finding the optimal threshold for SNNs for each of the $M$ predictions, considering class- and data-dependent adaptation at each epoch, would significantly increase training time. Therefore, using agreement-based pseudo-labeling without additional thresholds achieves optimal performance by leveraging sufficient and diverse predictions in our SNN setting.

\paragrapht{Impact of time steps and the number of collections.}
Additionally, we varied the number of time steps ($T$) and prediction collections ($M$). As shown in Table~\ref{tab:abl_timestep_collection}, increasing $T$ improves accuracy but results in higher energy consumption. Thus, we adopt $T=4$ as a trade-off. Using multiple collections $M=3$ consistently improves generalization over single-prediction pseudo-labeling, especially at lower $T$, demonstrating the strength of temporal diversity in SpikeMatch even under limited computation. Additional ablations with network variations and hyperparameter details are included in the Appendix.

\begin{table}[t!]
    \centering
    \resizebox{\columnwidth}{!}{
    \begin{tabular}{l|cc}
    \toprule[1pt]
        Ablation &CIFAR-10 &CIFAR-100  \\
        \cmidrule(lr){1-1} \cmidrule(lr){2-2} \cmidrule(lr){3-3}
        SpikeMatch &86.94 &48.31 \\
        \cmidrule(lr){1-1} \cmidrule(lr){2-2} \cmidrule(lr){3-3}
        - Averaged prediction &80.81 &44.05 \\
        - Intra prediction &80.46 &44.02 \\
        - w/o DA &76.32 &42.91 \\
        - w/ threshold &83.17 &41.73 \\
    \bottomrule[1pt]
    \end{tabular}}
    \caption{\textbf{Ablation studies}}
    \label{tab:abl_module}\vspace{-5pt}
\end{table}

\begin{table}[t]
    \centering
    \begin{tabular}{l|cc}
    \toprule[1pt]
        $T$  &$M$=1 &$M$=3\\
        \cmidrule(lr){1-1} \cmidrule(lr){2-2} \cmidrule(lr){3-3}
        4 &80.81 &86.94\\
        6 &81.47 &87.84\\
        8 &- &88.54 \\
        \bottomrule[1pt]
    \end{tabular}
    \caption{\textbf{Ablation studies of time step ($T$) and the number of collections ($M$)}}\vspace{-10pt}
    \label{tab:abl_timestep_collection}
\end{table}

\section{Conclusion}
\label{sec:conclusion}
In this work, we presented SpikeMatch, a semi-supervised learning (SSL) framework for spiking neural networks (SNNs) that utilize their temporal dynamics via the leakage factor to achieve diverse pseudo-labeling within a co-training framework. By utilizing agreement among multiple predictions from a single SNN, SpikeMatch produces reliable pseudo-labels from weakly augmented unlabeled samples to train on strongly augmented ones, effectively mitigating confirmation bias and enhancing feature learning with limited labeled data. Experiments show that SpikeMatch outperforms existing SSL methods adapted to SNN backbones across various benchmark.

{
    \small
    \bibliographystyle{ieeenat_fullname}
    \bibliography{main}
}
\clearpage
\renewcommand{\thesection}{\Alph{section}}
\renewcommand{\thefigure}{A.\arabic{figure}}
\renewcommand{\thetable}{A.\arabic{table}}

\setcounter{section}{0}
\setcounter{figure}{0}
\setcounter{table}{0}
\maketitlesupplementary

\section{Experimental Setup}
\subsection{Semi-supervised training details}
For each dataset CIFAR~\cite{krizhevsky2009learning}, STL-10~\cite{coates2011analysis}, and ImageNet~\cite{deng2009imagenet}, we select a various number of labeled data, sampling an equal number of examples from each class. In all datasets, we randomly sample a small subset as labeled data and treat the remaining examples as unlabeled, except for STL-10, which originally includes an unlabeled dataset. The hyperparameter configurations for semi-supervised learning are summarized in Table~\ref{sup_tab:semi_setting}, where we generally follow the standard semi-supervised learning protocol~\cite{sohn2020fixmatch, chen2023softmatch}, with certain adjustments tailored to the SNN backbone.

\begin{table*}[h]
\centering
\begin{tabular}{l|c c c c}
 \toprule[1pt]
 Dataset & CIFAR-10 & CIFAR-100 & STL-10 & ImageNet \\
\cmidrule(lr){1-1} \cmidrule(lr){2-2} \cmidrule{3-3} \cmidrule(lr){4-4} \cmidrule(lr){5-5}
\multirow{2}{*}{Model} & \multirow{2}{*}{VGG9} & \multirow{2}{*}{VGG11} & VGG11 & \multirow{2}{*}{SEW-ResNet-50} \\
 &  &  & SEW-ResNet-34 &  \\
\cmidrule(lr){1-1} \cmidrule(lr){2-2} \cmidrule(lr){3-3} \cmidrule(lr){4-4} \cmidrule(lr){5-5}
Weight decay & \multicolumn{3}{c}{$5 \times 10^{-4}$}  & $3 \times 10^{-4}$ \\
\cmidrule(lr){1-1} \cmidrule(lr){2-4} \cmidrule(lr){5-5}
Iterations & \multicolumn{3}{c}{$2^{18}$} & $2^{20}$ \\
\cmidrule(lr){1-1} \cmidrule(lr){2-4} \cmidrule(lr){5-5}
Batch size & \multicolumn{3}{c}{32} & 16 \\
\cmidrule(lr){1-1} \cmidrule(lr){2-4} \cmidrule(lr){5-5}
label unlabel ratio ($\mu$) & \multicolumn{3}{c}{7} & 1 \\
\cmidrule(lr){1-1} \cmidrule(lr){2-4} \cmidrule(lr){5-5}
Learning rate & \multicolumn{4}{c}{0.03} \\
\cmidrule(lr){1-1} \cmidrule(lr){2-5}
SGD momentum & \multicolumn{4}{c}{0.9} \\
\cmidrule(lr){1-1} \cmidrule(lr){2-5}
EMA decay & \multicolumn{4}{c}{0.999} \\
\cmidrule(lr){1-1} \cmidrule(lr){2-5}
Weak Augmentation & \multicolumn{4}{c}{Random Crop, Random Horizontal Flip} \\
\cmidrule(lr){1-1} \cmidrule(lr){2-5}
Strong Augmentation & \multicolumn{4}{c}{RandAugment~\cite{cubuk2020randaugment}} \\
\bottomrule[1pt]
\end{tabular}
\caption{\textbf{Hyperparameter setting for semi-supervised learning.}}
\label{sup_tab:semi_setting}
\end{table*}

\subsection{Spiking-specific training}
To address the non-differentiability of spike generation, we adopt a triangular-shaped surrogate gradient function~\cite{deng2022temporal}.
As we utilize the leaky integrate-and-fire (LIF) model~\cite{burkitt2006review}, for the direct training of spiking neural networks (SNNs), we use the spatio-temporal back-propagation (STBP) rule for gradient descent training~\cite{wu2018spatio}. The membrane potential before spiking at time step $t \in \{1, ..., T\}$ and layer $l \in \{1, ..., L\}$ denoted as ${u}^{l,pre}(t)$, is updated based on the pre-synaptic input  $W^l s^{l-1}(t)$, which is calculated by multiplying the synaptic weight ${W^l}$ by the input at time step  $t$, ${s}^{l-1}(t)$:

\begin{equation}
    {u}^{l,pre}(t) = \tau{u}^{l}(t-1) + W^l s^{l-1}(t),
\end{equation}
where $\tau$ is the leakage factor and it balances the contribution of membrane potential after spiking at previous time step, ${u}^{l,pre}(t)$.  
The neuron generates spike output $s^l(t)$, when the accumulated membrane potential ${u}^{l,pre}(t)$ exceeds the pre-defined firing threshold $V_{th}$, which we set it to 1:
\begin{equation}
    {s}^{l}(t) = \Theta({u}^{l,pre}(t) - V_{th}) = 
    \begin{cases}
    1,\quad if \ {u}^{l,pre}(t) \geq V_{th} \\
    0,\quad otherwise
    \end{cases}, 
\end{equation}
\begin{equation}
    {u}^{l}(t) = {u}^{l,pre}(t)(1-{s}^{l}(t)),
\end{equation}
where $\Theta$ represents the Heaviside step function, and ${u}^{l}(t)$ is reset to zero (hard reset) if ${u}^{l,pre}(t)$ exceed the threshold $V_{th}$. In SNNs, this process is applied iteratively. For the gradient descent training via back-propagation, the gradient calculation with the cross-entrophy loss $L$ under the STBP rule can be expressed as: 
\begin{equation}
            \frac{\partial L}{\partial {W}^l} = 
            \sum_t
            \frac{\partial L}{\partial {s}^{l}(t)}
            \frac{\partial {s}^{l}(t)}{\partial {u}^{l,pre}(t)}
            \frac{\partial {u}^{l,pre}(t)}{\partial {W}^l},
\end{equation}
where $\frac{\partial {s}^{l}(t)}{\partial {u}^{l,pre}(t)}$ is non-differentiable, we use a surrogate gradient as an approximate derivative of the spike activity. In this paper, we use a triangular function as the surrogate gradient~\cite{deng2022temporal}:
\begin{equation}
    \frac{\partial {s}^{l}(t)}{\partial {u}^{l,pre}(t)} = \frac{1}{\gamma^2} \max\left(0, \gamma - \lvert {u}^{l,pre}(t) - V_{th} \rvert \right).
\end{equation} 
where $\gamma$ is a hyper-parameter, which we set to 1.

\subsection{Energy estimation}

For the estimation of energy consumption, we follow the formulation proposed by Kundu et al.~\cite{kundu2021hire}, based on the direct input coding scheme. The total inference energy of an L-layer SNN is computed as

\begin{equation}
\mathcal{E}_{\text{SNN}}^{\text{direct}} = \mathcal{F}_1 \cdot E_{\text{MAC}} + \sum_{l=2}^{L} \mathcal{F}_l \cdot E_{\text{AC}},
\end{equation}

where $\mathcal{F}_l$ denotes the number of operations in layer l, and $E_{\text{MAC}}$ and $E_{\text{AC}}$ represent the energy cost of multiply-accumulate and accumulate operations, respectively. The first layer operates on real-valued inputs and thus incurs MAC operations, while all subsequent layers process binary spikes and use AC operations.

The operation count $\mathcal{F}_l$ for a convolutional layer is calculated as
\begin{equation}
\mathcal{F}_l = (k^l)^2 \cdot H^l \cdot W^l \cdot C^l_{\text{in}} \cdot C^l_{\text{out}} \cdot \zeta^l,
\end{equation}

where $k^l$ is the kernel size, $H^l$ and $W^l$ are the height and width of the output feature map, $C^l_{\text{in}}$ and $C^l_{\text{out}}$ denote the number of input and output channels, and $\zeta^l \in [0, 1]$ is the average spiking activity at layer $l$. This activity reflects the temporal sparsity in SNNs and scales the effective number of computations accordingly.

Following~\cite{kundu2021hire}, we assume an energy cost of $E_{\text{MAC}} = 4.6\,\text{pJ}$ for MAC operations and $E_{\text{AC}} = 0.9\,\text{pJ}$ for AC operations under 32-bit floating point (FP) precision. 

\section{Additional Results}

\begin{table}[t]
    \centering
    \begin{tabular}{c|c}
    \toprule[1pt]
          &DVS-CIFAR10 \\
        \cmidrule(lr){1-1} \cmidrule(lr){2-2}
        SoftMatch &36.40  \\
        FreeMatch &37.40  \\
        \hline
        SpikeMatch &{\textbf{49.80}} \\
         \bottomrule[1pt]
    \end{tabular}
    \caption{\textbf{Accuracy (\%) on DVS-CIFAR10 with 1\% labels.}}\vspace{-15pt}
    \label{sup_tab:tab_dvs}
\end{table}
\subsection{Results on CIFAR-10-DVS}
To validate the effectiveness of SpikeMatch beyond normal datasets, we conduct additional experiments on CIFAR10-DVS using event-based data. For strong augmentation, we adopt EventAugment~\cite{gu2024eventaugment}, which avoids image-specific transformations (e.g., brightness or color jitter) that are unsuitable for DVS data, while preserving the temporal structure of events. As shown in Table~\ref{sup_tab:tab_dvs}, SpikeMatch outperforms SoftMatch~\cite{chen2023softmatch} and FreeMatch~\cite{wang2022freematch} even on this neuromorphic dataset, supporting that our multi-view agreement remains effective when the temporal diversity arises from the data itself.

\section{Extended Analysis}
\begin{table}[t]
\begin{center}
\resizebox{\columnwidth}{!}
{
    \begin{tabular}{lcc|c}
    \toprule[1pt]
    Dataset &Network &Method &\textbf{Accuracy} \\
    \cmidrule(lr){1-1} \cmidrule(lr){2-2} \cmidrule(lr){3-3} \cmidrule(lr){4-4}
    \multirow{5}{*}{CIFAR-10}
    &\multirow{5}{*}{WRN-28-2}
    &FixMatch &87.98 \\
    & &AdaMatch &88.20 \\
    & &FreeMatch &88.80 \\
    & &SoftMatch &89.19 \\
    \cmidrule(lr){3-3} \cmidrule(lr){4-4}
    & &Ours &\textbf{90.04} \\
    \hline
    \multirow{5}{*}{CIFAR-100}& \multirow{5}{*}{WRN-28-8}
    &FixMatch &55.22 \\
    & &AdaMatch &56.83 \\
    & &FreeMatch &57.62 \\
    & &SoftMatch &57.38    \\
    \cmidrule(lr){3-3} \cmidrule(lr){4-4}
    & &Ours &\textbf{60.84} \\
    \bottomrule[1pt]      
    \end{tabular}
}
    \caption{\textbf{Accuracy (\%) using WRN with 4 time steps on CIFAR-10 using 400 labels per class and CIFAR-100 using 100 labels per class.}}\label{sup_tab:abl_network}
\end{center}
\end{table}
\subsection{Network configuration}
By default, we use the VGG network~\cite{sengupta2019going, deng2022temporal} in our experiments, as it is commonly used in other SNN methods. In contrast, SSL methods in ANNs typically use Wide ResNet (WRN)~\cite{zagoruyko2016wide} as the backbone. To demonstrate that our method performs well across different network architectures, we also conduct experiments using WRN on CIFAR-10 and CIFAR-100, utilizing 400 and 100 labeled samples per class, respectively.
As reported in Table~\ref{sup_tab:abl_network}, our method achieves the highest performance on WRN for both CIFAR-10 and CIFAR-100. 
These results demonstrate that our agreement-based pseudo-labeling method, which leverages the temporal dynamics of SNNs, is effective for semi-supervised learning in SNNs across various network architectures.

\label{sec:extended_exp}
\begin{table}[t!]
    \centering
    \resizebox{\columnwidth}{!}
    {
    \begin{tabular}{l|cccc}
    \toprule[1pt]
        Seed &0 &1 &2 &Mean  \\
        \cmidrule(lr){1-1} \cmidrule(lr){2-2} \cmidrule(lr){3-3} \cmidrule(lr){4-4} \cmidrule(lr){5-5}
        SoftMatch &83.65 &83.63 &83.55 &83.61$_{\pm 0.05}$ \\
        SpikeMatch &83.76 &85.35 &85.55 &\textbf{84.88$_{\pm 0.98}$} \\
         \bottomrule[1pt]
    \end{tabular}
    }
    \caption{\textbf{Accuracy (\%) on STL-10 with 250 labeled samples.}}
    \label{sup_tab:stl_sew}
\end{table}

\subsection{Results on STL-10}
To demonstrate the backbone-agnostic effectiveness of our method, we evaluate it on STL-10, a more complex dataset with high-resolution images, using 250 labeled samples. We conduct experiments with SEW-ResNet~\cite{fang2021deep}, a commonly used residual connection architecture for deep SNNs. For the SEW-ResNet backbone, we use 4 time steps, a triangular function as the surrogate gradient, and a hard reset mechanism after spiking, implemented by modifying the code provided by the authors. As shown in Table \ref{sup_tab:stl_sew}, our method achieves higher accuracy compared to the recent SSL method, SoftMatch~\cite{chen2023softmatch}. Even with a deeper network, our method achieves higher accuracy across all seed settings by effectively utilizing the temporal dynamics of SNNs for generalizable and reliable SSL training.

\begin{table}[t!]
    \centering
\resizebox{\columnwidth}{!}
    {
    \begin{tabular}{lc|c}
    \toprule[1pt]
    Reset\&SG &Method &Accuracy \\
    \cmidrule(lr){1-1} \cmidrule(lr){2-2} \cmidrule(lr){3-3}
   \multirow{8}{*}{\makecell{Soft\\\&Rectangular}}
   &UDA~\cite{xie2020unsupervised} &74.26  \\
   &FixMatch~\cite{sohn2020fixmatch} &57.62  \\
   &FlexMatch~\cite{zhang2021flexmatch} &66.94  \\
       &AdaMatch~\cite{berthelot2021adamatch} &88.58 \\
       &FreeMatch~\cite{wang2022freematch} &89.49 \\
       &SoftMatch~\cite{chen2023softmatch} &87.49 \\
       &SequenceMatch~\cite{nguyen2024sequencematch} &76.88 \\
       \cmidrule(lr){2-2} \cmidrule(lr){3-3}
       &SpikeMatch &\textbf{89.73} \\
   \bottomrule[1pt]
    \end{tabular}
    }
    \caption{\textbf{Accuracy (\%) with soft reset and rectangular surrogate gradient function (SG) on CIFAR-10 using 4 labels per class with a time step 8.}}
    \label{sup_tab:tab_abl_lif}
\end{table}
\begin{table}[t]
    \centering
\resizebox{\columnwidth}{!}
    {
    \begin{tabular}{c|ccccc}
    \toprule[1pt]
          &$\bm{T=4}$ &$T=4$ &$T=3$ &$T=2$ &\multirow{2}{*}{$T=1$} \\
          &$\bm{(M=3)}$ &$(M=2)$ &$(M=3)$ &$(M=2)$ & \\
        \hline
        Acc.(\%) &{\textbf{88.13}} &85.97 &87.85 &84.03 &64.38 \\
         \bottomrule[1pt]
    \end{tabular}
    }
    \caption{\textbf{Performance across time steps ($T$) and groups ($M$).}}
    \label{sup_tab:tab_t_m}
\end{table}
\subsection{LIF configuration}
We also conducted ablation experiments on different configurations of the Leaky Integrate-and-Fire (LIF) model. By default, we used a hard reset mechanism with a triangular function as the surrogate gradient to handle the non-differentiability in direct training of SNNs, and we set the number of time steps  $T = 4$. In the ablation study, we employed a soft reset mechanism and a rectangular function~\cite{wu2018spatio} as the surrogate gradient, increasing the time steps $T$ to 8. With  $T = 8$, we partitioned the outputs into collections of sizes $\{3, 2, 3\}$.

As shown in Table~\ref{sup_tab:tab_abl_lif}, by increasing the time steps and using a soft reset, which is known for higher information preservation, we achieved higher accuracy on CIFAR-10 with 4 labeled examples per class. These results demonstrate that our method can be extended to longer time steps and various network configurations in SNNs while still outperforming other SSL methods in ANNs. As shown on Table~\ref{sup_tab:tab_t_m}, while longer time steps contribute to better feature learning, even with the same $T$, treating temporally grouped predictions as distinct views in our agreement-based framework yields more substantial performance gains. These results demonstrate that temporal dynamics are fundamental to learning discriminative features in SNNs, and that leveraging them as diverse views through agreement is especially beneficial when labeled data is limited. The noticeable performance drop when the time step is reduced to one highlights that, in the context of semi-supervised learning with scarce labels, exploiting temporal diversity is particularly important for enabling reliable pseudo-labeling.

\begin{table}[!t]
    \centering
    \begin{tabular}{l|ccc}
    \toprule[1pt]
          &$\bm{n=3}$ &$n=4$ &$n=5$ \\
        \cmidrule(lr){1-1} \cmidrule(lr){2-2} \cmidrule(lr){3-3} \cmidrule(lr){4-4}
        Acc.(\%) &{\textbf{88.13}} &87.15 &86.95 \\
         \bottomrule[1pt]
    \end{tabular}
    \caption{\textbf{Ablation study on number of RandAugment Transforms.}}
    \label{sup_tab:tab_abl_randaug}
\end{table}
\begin{table}[]
    \centering
    \begin{tabular}{l|ccccc}
    \toprule[1pt]
          &$\lambda=0.5$ &$\bm{\lambda=1.}$ &$\lambda=2.$ &$\lambda=4.$ \\
        \cmidrule(lr){1-1} \cmidrule(lr){2-2} \cmidrule(lr){3-3} \cmidrule(lr){4-4} \cmidrule(lr){5-5} 
        Acc.(\%) &87.04 &{\textbf{88.13}} &88.04 &87.71 \\
         \bottomrule[1pt]
    \end{tabular}
    \caption{\textbf{Ablation study on unlabeled loss weight}}
    \label{sup_tab:abl_lambda}
\end{table}
\subsection{Sensitivity analysis}
In our experiments, we use random cropping and random horizontal flipping as weak augmentations, and adopt RandAugment~\cite{cubuk2020randaugment} as a strong augmentation. We conduct ablation studies on both the augmentation strategy and the loss weight $\lambda$, which balances the supervised and unsupervised objectives. As shown in Table~\ref{sup_tab:tab_abl_randaug}, varying the number of RandAugment transformations exhibits trends consistent with ANN-based SSL methods such as UDA~\cite{xie2020unsupervised} and FixMatch~\cite{sohn2020fixmatch}, leading us to set it to 1 by default. Regarding $\lambda$, we observe that performance remains stable across a wide range, confirming that our default choice of assigning equal weight to the supervised and unsupervised losses is near-optimal, as shown in Table~\ref{sup_tab:abl_lambda}.


\begin{figure}[t!]
\centering	{\includegraphics[width=\linewidth]{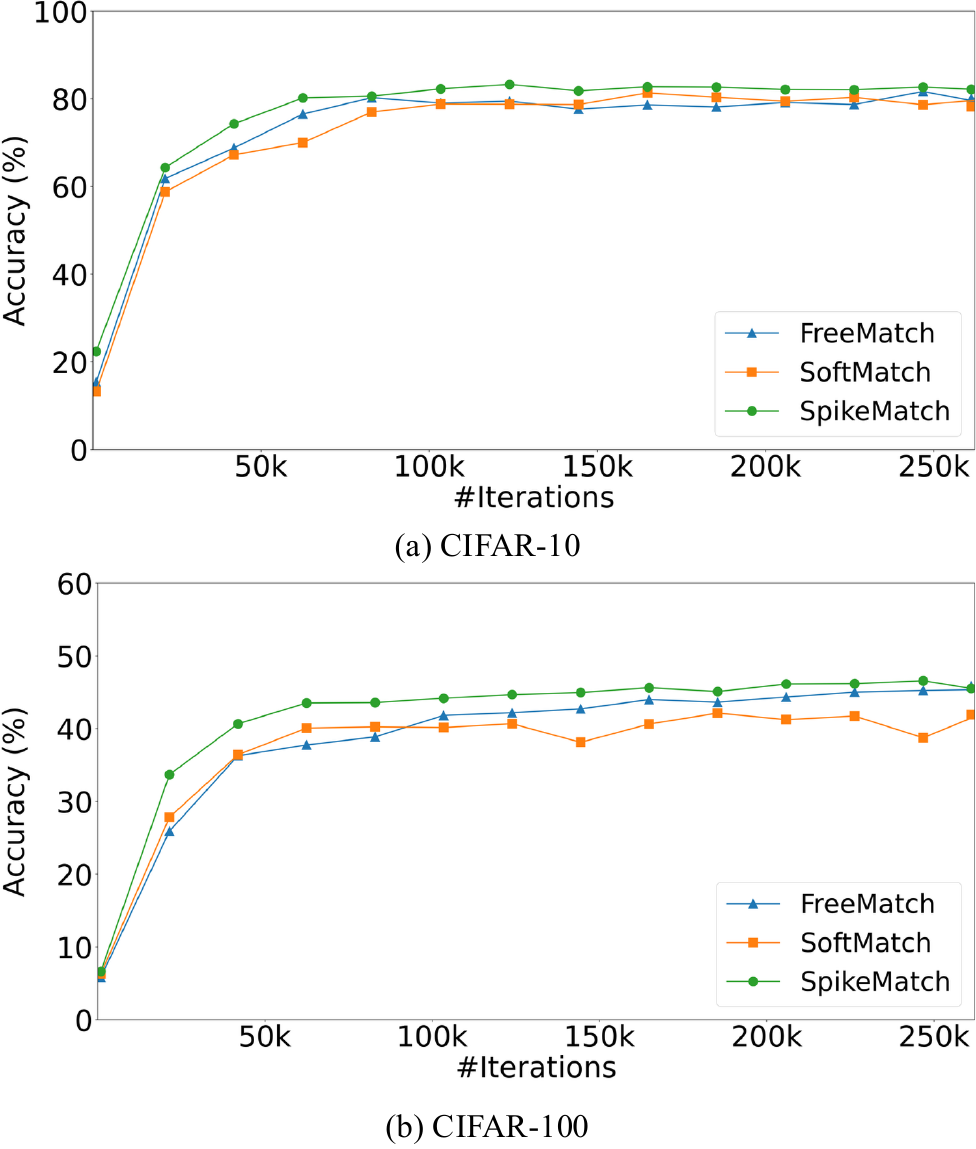}}
    \caption{\textbf{Convergence speed of proposed method with SoftMatch~\cite{chen2023softmatch} and FreeMatch~\cite{wang2022freematch} on (a) CIFAR-10 and (b) CIFAR-100 with 4 labeled samples per class.}}
    \label{sup_fig:convergence_speed}
\end{figure}
\begin{figure}[ht]
\centering
	\centering	{\includegraphics[width=\linewidth]{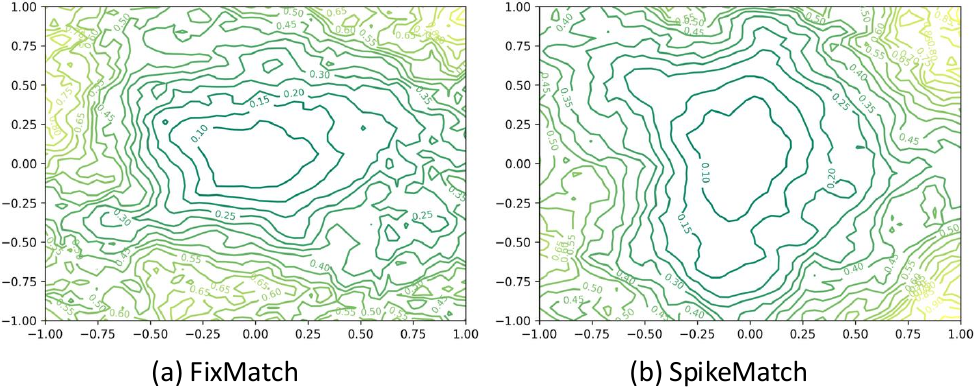}}
    \caption{\textbf{Loss landscape of (a) FixMatch and (b) SpikeMatch of labeled data trained on CIFAR-10 with 250 labeled samples.}}
    \label{sup_fig:loss_landscape}
\end{figure}
\begin{figure}[h]
    {\includegraphics[width=\columnwidth]{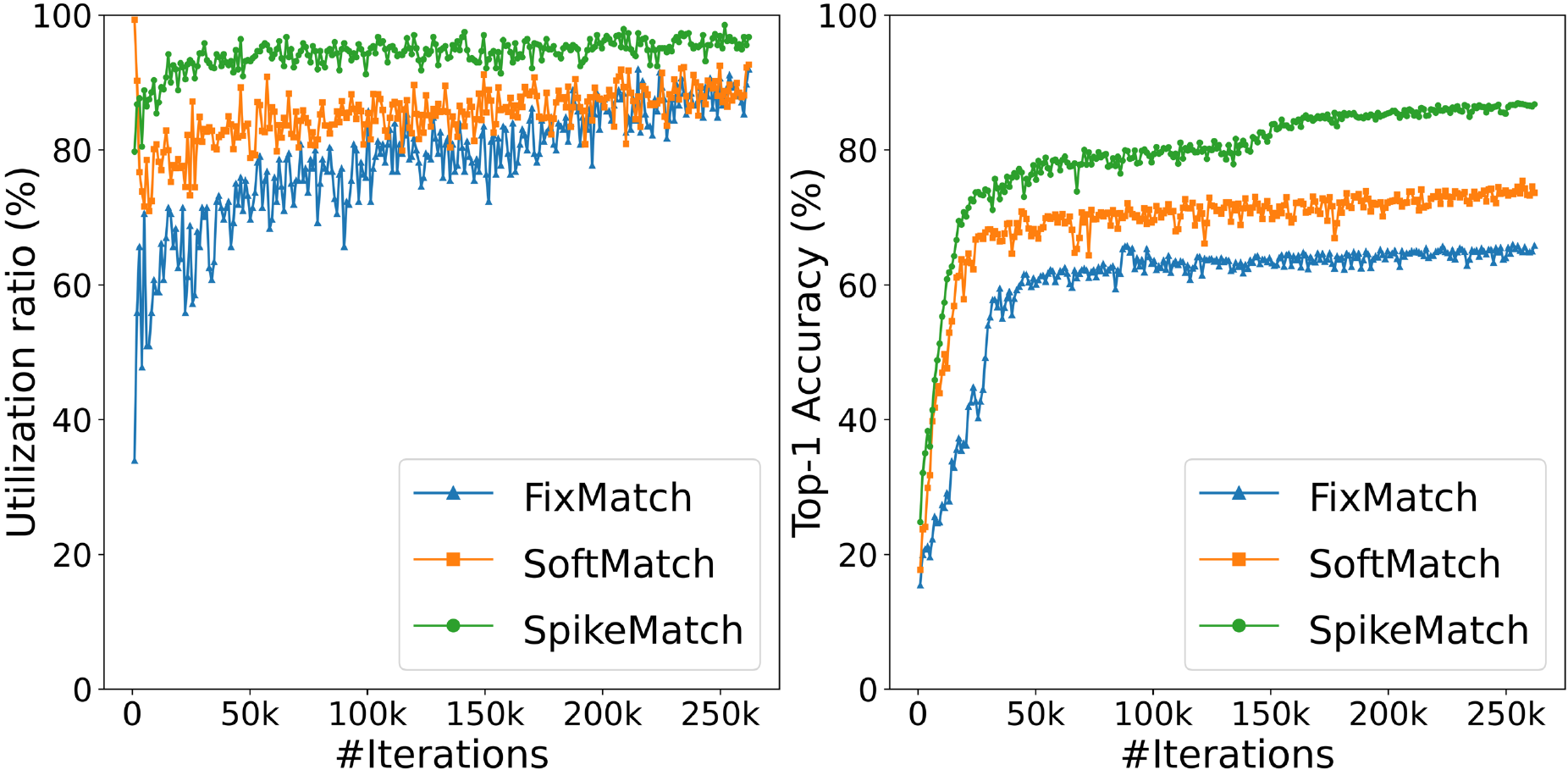}}
    \caption{\textbf{Utilization graphs (Left) comparing SpikeMatch, FixMatch, and SoftMatch on CIFAR-10 with 4 labeled samples per class and accuracy graph (Right).}}
    \label{sup_fig:util_ratio}
\end{figure}

\begin{figure}[!ht]
    \centering	{\includegraphics[width=0.9\columnwidth]{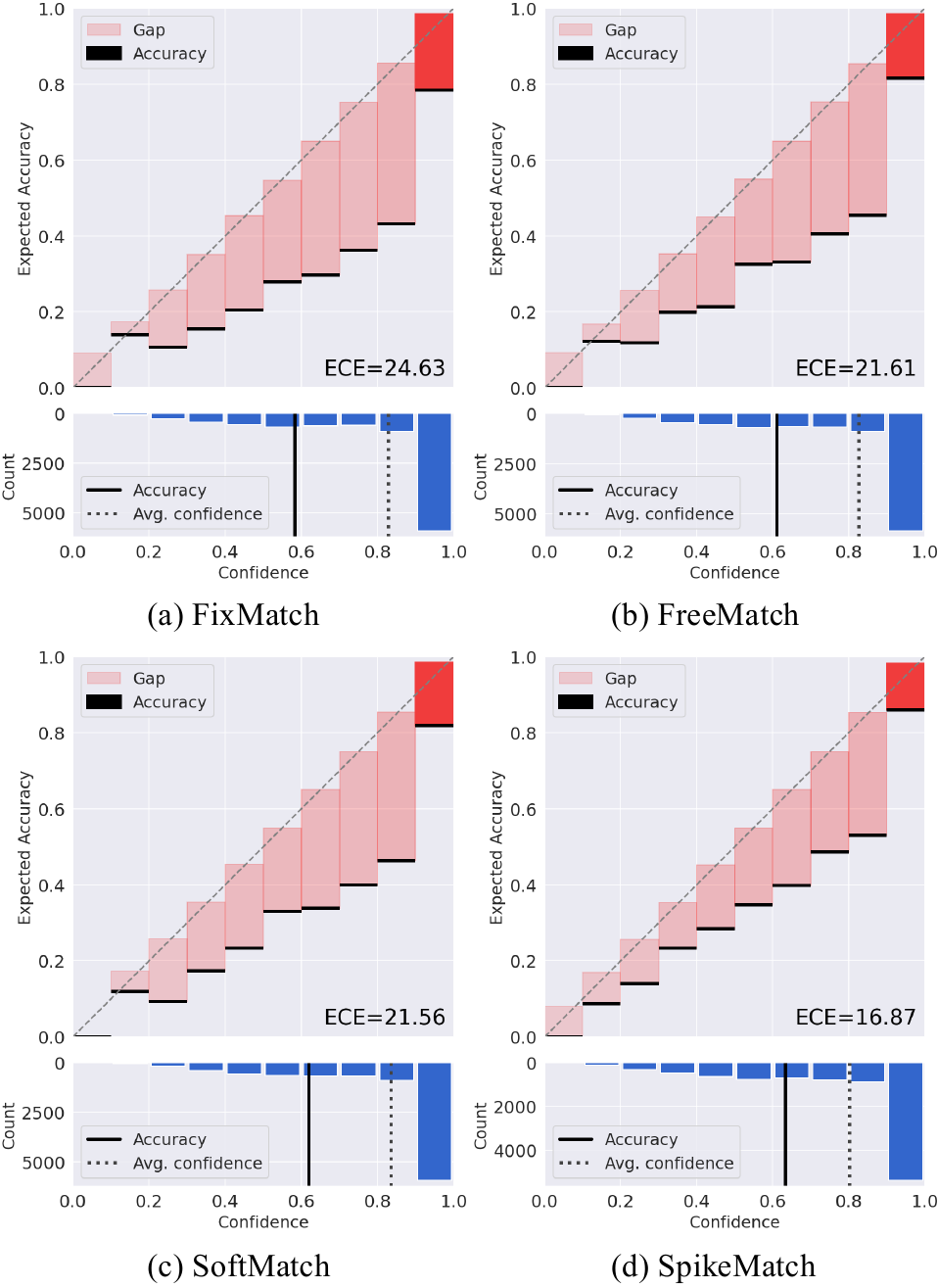}}
    \caption{\textbf{Reliability diagram (top) and confidence histogram (bottom).} (a) FixMatch, (b) FreeMatch, (c) SoftMatch, and (d) SpikeMatch are trained on CIFAR-100 with 2500 labeled samples and tested using 10 interval bins.}
    \label{sup_fig:ece_calibration}
\end{figure}
\begin{figure*}[hbt]
	\centering	{\includegraphics[width=0.8\linewidth]{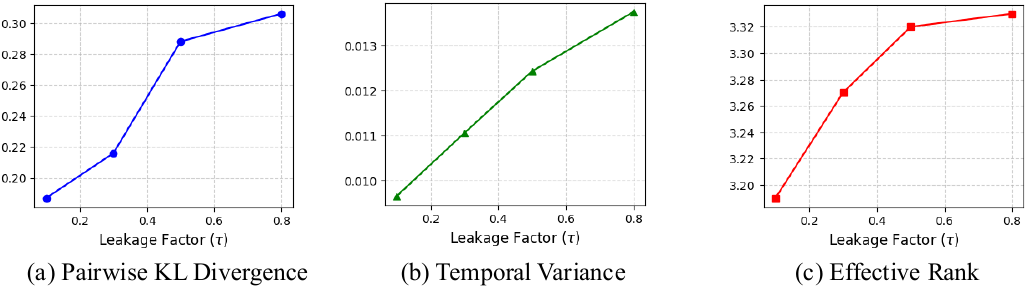}}
 	\caption{\textbf{(a) Pairwise KL divergence, (b) Temporal Variance, and (c) Effective Rank at varying leakage factors ($\tau$) on CIFAR-10 with 4 time steps.}}
	\label{sup_fig:kl_var_erank}
\end{figure*}

\subsection{Convergence speed}
We compare the convergence speed of SpikeMatch with SoftMatch~\cite{chen2023softmatch} and FreeMatch~\cite{wang2022freematch} on CIFAR-10 with 4 labeled samples per class. As shown in Figure~\ref{sup_fig:convergence_speed}, SpikeMatch requires fewer iterations to converge while achieving higher accuracy by leveraging temporal efficient training~\cite{deng2022temporal} and effectively utilizing abundant unlabeled data with diverse pseudo-labels.
 
\subsection{Loss landscape}
In SSL, labeled data are typically much smaller in quantity than unlabeled data, resulting in poor generalization during labeled data training~\cite{huang2023flatmatch}. To address this, we utilize Temporal Efficient Training (TET) loss~\cite{deng2022temporal} to enhance generalization without requiring additional training to find flatter minima in the context of limited labeled samples. As shown in Figure~\ref{sup_fig:loss_landscape}, SpikeMatch using TET loss achieves a smoother loss landscape compared to FixMatch~\cite{sohn2020fixmatch}, which uses averaged temporal outputs as prediction. This suggests that the low generalization issue associated with small labeled datasets in SSL can be moderately alleviated by leveraging the inherent characteristics of SNNs without relying on pre-training methods.

\subsection{Pseudo-label utilization ratio}
  The utilization ratio indicates the number of pseudo-labels used for training: for SpikeMatch, those in agreement; for FixMatch and SoftMatch, those surpassing the confidence threshold. As shown in Figure~\ref{sup_fig:util_ratio}, with our agreement method, unreliable predictions are filtered out in early training, effectively reducing confirmation bias. The utilization ratio increases steadily throughout training, indicating both the stability and high quantity of reliable pseudo-labels.

\subsection{Expected Calibration Error (ECE)}
Model calibration is crucial in semi-supervised learning (SSL) to mitigate confirmation bias~\cite{chen2022semi, nguyen2023boosting, nguyen2024sequencematch, loh2022importance}. To demonstrate the effectiveness of SpikeMatch, we compare the calibration of FixMatch~\cite{sohn2020fixmatch}, FreeMatch~\cite{wang2022freematch}, and SoftMatch~\cite{chen2023softmatch} with our approach using Expected Calibration Error (ECE), confidence histograms, and reliability diagrams to visualize calibration error. In this work, we compute ECE as the
expected difference between the accuracy and confidence across 10 bins for all samples. A high ECE value indicates that the model is over-confident, which can lead to unreliable pseudo-labels and confirmation bias during training. Lower ECE values indicate better calibration.

As shown in Figure~\ref{sup_fig:ece_calibration}, SpikeMatch achieves the best calibration among all methods. Its reliability diagram aligns most closely with the ideal diagonal, and its confidence histogram indicates reduced over-confidence. The red bars show the gap between confidence and accuracy in each bin, demonstrating its effectiveness in addressing confirmation bias.


\section{Empirical Insights}
\subsection{Leakage factor and temporal diversity}

To support our claim that a higher leakage factor $\tau$ enhances temporal diversity in spiking neural networks, we conduct a supervised analysis using multiple diversity metrics. While cosine similarity is used in the main paper to measure differences between temporal predictions, we extend the analysis by incorporating additional metrics that capture different aspects of diversity. We use pairwise KL divergence to evaluate the dissimilarity of predictions over time, temporal variance to measure the dispersion of features, and effective rank to quantify the dimensional richness of temporal representations.

We vary $\tau$ from 0.1 to 0.8 and evaluate these metrics on the CIFAR-10 dataset. As shown in Figure~\ref{sup_fig:kl_var_erank}, all three metrics increase consistently with larger $\tau$. In (a), KL divergence grows as predictions across time steps become more distinct. In (b), temporal variance increases, indicating that features are more widely distributed. In (c), effective rank rises, showing that the temporal features span a higher-dimensional space with reduced redundancy.

These results provide strong empirical evidence from multiple perspectives that a higher leakage factor naturally induces temporal diversity. This property plays an important role in the co-training-inspired design of SpikeMatch.

\section{Visualization on CIFAR-10 and STL-10.}
\begin{figure*}[!t]
    \centering
        \includegraphics[width=0.9\linewidth]{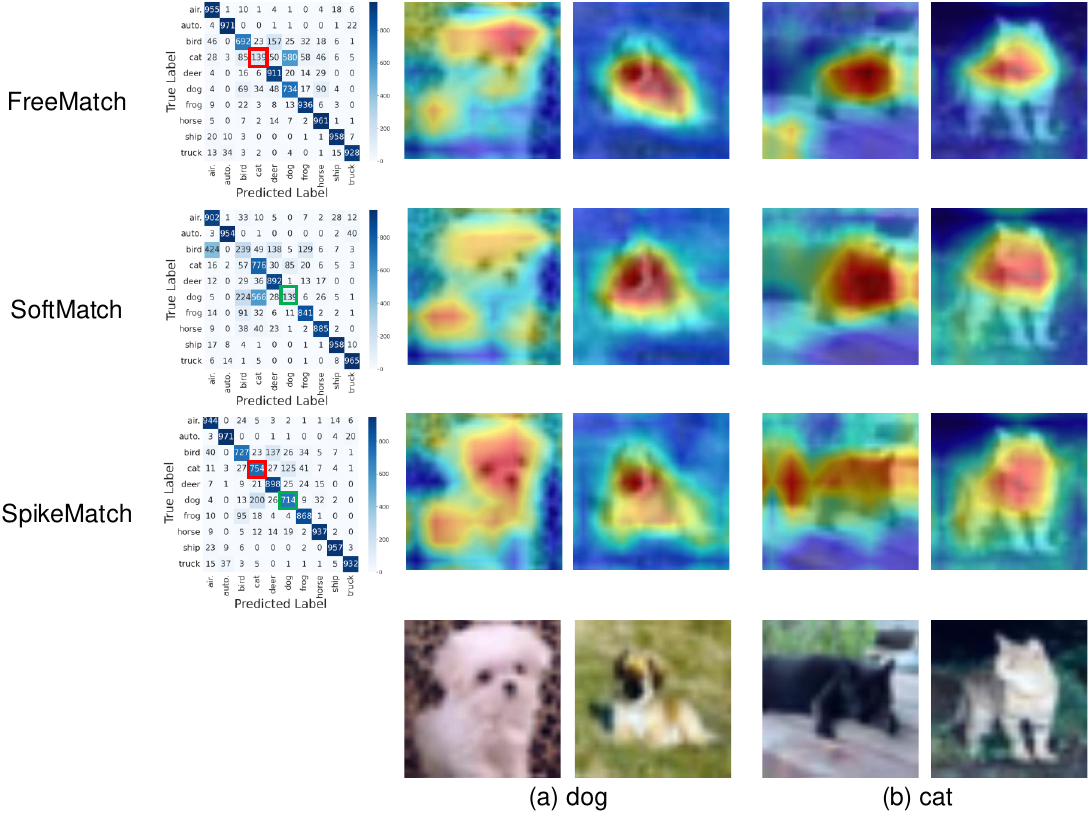}
    \hfill
 	\caption{\textbf{Confusion matrix and spiking attention map (SAM)~\cite{kim2021visual} on CIFAR-10.} Unlike FreeMatch (\textbf{first row}) and SoftMatch (\textbf{second row}), which struggle with the cat and dog classes, our method (\textbf{last row}) captures more distinctive features of (a) dog and (b) cat. This leads to a more diagonal confusion matrix, reducing confusion among similar classes.}
	\label{sup_fig:confusion_cifar}
\end{figure*}

\begin{figure*}[!t]
    \centering
        \includegraphics[width=0.9\linewidth]{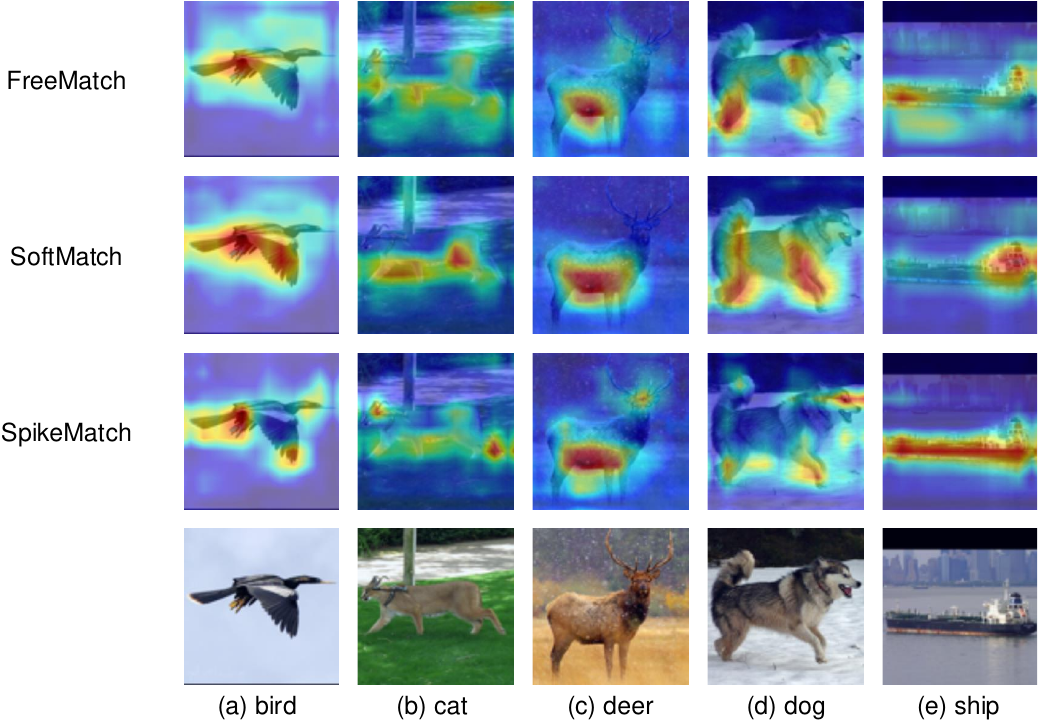}
    \hfill
 	\caption{\textbf{Spiking attention map (SAM)~\cite{kim2021visual} on STL-10.} Unlike FreeMatch (\textbf{first row}) and SoftMatch (\textbf{second row}), our method (\textbf{last row}) more effectively captures multiple discriminative features through the agreement-based approach.}
	\label{sup_fig:sam_stl10}
\end{figure*}

As additional examples, Figure~\ref{sup_fig:confusion_cifar} and Figure~\ref{sup_fig:sam_stl10} illustrate how our method captures multiple discriminative features by leveraging the inherent temporal diversity of SNNs. Specifically, our agreement-based temporal dynamic framework enhances class distinction by extracting diverse and complementary features across time steps. These qualitative results further support the conclusion that our approach remains effective even under severe label constraints.

\clearpage

\clearpage
\FloatBarrier


\end{document}